%% file: xiaomi.tex
\documentclass[11pt, a4paper, logo, copyright, nonumbering]{xiaomi}

\usepackage[authoryear, sort&compress, round]{natbib}
\usepackage{dblfloatfix}
\usepackage{ulem}
\usepackage{caption}
\usepackage{dramatist}
\usepackage{xspace}
\usepackage{pifont}
\usepackage{multirow}
\usepackage{tcolorbox}
\usepackage{xltabular}
\usepackage{longtable}
\usepackage{hyperref}
\usepackage{wrapfig}
\usepackage{graphicx}

\usepackage{amsfonts}
\usepackage{amsmath}
\usepackage{amssymb}
\usepackage{lineno}
\usepackage{multirow}
\usepackage{adjustbox}

\usepackage[bottom]{footmisc}
\usepackage{algorithm}
\usepackage{algpseudocode}
\usepackage{CJKutf8}
\usepackage{subcaption}
\usepackage{setspace}
\usepackage{makecell}
\usepackage{graphicx}
\usepackage{multicol}
\usepackage{xspace}
\newcommand{\aname}{HySparse\xspace}  %Hybrid SA

\usepackage{cleveref}

\algnewcommand{\Initialize}{\textbf{Initialize: }}

\defcitealias{nondeter}{T. M. Lab}

\definecolor{xiaomiorange}{HTML}{FF6901}

\begin{abstract}
    \input{00-abs}
\end{abstract}

\begin{document}

% {
%     \bgroup
%     \setlength{\parindent}{0pt}
%     \vspace*{3pt} 
%     \begin{adjustwidth}{0pt}{0pt}  
%     \begin{center} 
%     {\titlefont HySparse: A Hybrid Sparse Attention Architecture \\ with Oracle Token Selection and KV Cache Sharing \par}
%     {
%     \vskip5pt
%     {\normalfont\sffamily\fontsize{11}{15}\selectfont Wenhan Ma$^{\dagger\ddagger*}$ ~~~~~Hailin Zhang$^{\ddagger}$ ~~~~~Liang Zhao$^{\ddagger}$ ~~~~~Yifan Song$^{\dagger\ddagger}$} \\
%     {\normalfont\sffamily\fontsize{11}{15}\selectfont Yudong Wang$^{\dagger\ddagger}$ ~~~~~Zhifang Sui$^{\dagger\diamond}$ ~~~~~Fuli Luo$^{\S\diamond}$} \\
%     \vskip10pt
%     {\normalfont\sffamily\fontsize{11}{15}\selectfont $^{\dagger}$State Key Laboratory of Multimedia Information Processing, School of } \\
%     {\normalfont\sffamily\fontsize{11}{15}\selectfont  Computer Science, Peking University} \\
%     {\normalfont\sffamily\fontsize{11}{15}\selectfont $^{\ddagger}$LLM-Core Xiaomi} \\
%     {\normalfont\sffamily\fontsize{11}{15}\selectfont $^{\S}$Independent Researcher}
%     \vskip10pt
%     }
%     \end{center}
%     \end{adjustwidth}
%     \egroup
%     {\abscontent}
%     \thispagestyle{firststyle} 
% }

% \renewcommand{\thefootnote}{\fnsymbol{footnote}}
% \footnotetext[0]{$^{*}$Work done during internship at Xiaomi Corporation.}
% \footnotetext[0]{$^{\diamond}$Co-corresponding authors.}
% \renewcommand{\thefootnote}{\arabic{footnote}}

{
\bgroup
\setlength{\parindent}{0pt}
\vspace*{3pt}
\begin{adjustwidth}{0pt}{0pt}
\begin{center}
{\titlefont HySparse: A Hybrid Sparse Attention Architecture \\ with Oracle Token Selection and KV Cache Sharing \par}
{
\vskip5pt
{\normalfont\sffamily\fontsize{11}{15}\selectfont
Yizhao Gao ~~~  %$^{*}$
Jianyu Wei ~~~
Qihao Zhang ~~~
Yu Cheng ~~~
Shimao Chen \\
Zhengju Tang ~~~
Zihan Jiang ~~~
{\normalfont\sffamily\fontsize{11}{15}\selectfont
Yifan Song ~~~
Hailin Zhang ~~~
Liang Zhao \\
Bo Yang ~~~
Gang Wang ~~~
Shijie Cao ~~~
Fuli Luo$^{\diamond}$}
\\
\vskip10pt
{\normalfont\sffamily\fontsize{11}{15}\selectfont LLM-Core Xiaomi}
\vskip10pt
}
}
\end{center}
\end{adjustwidth}
\egroup
{\abscontent}
\thispagestyle{firststyle}
}

\renewcommand{\thefootnote}{\fnsymbol{footnote}}
%\footnotetext[0]{$^{*}$Work done during internship at Xiaomi Corporation.}
\footnotetext[0]{$^{\diamond}$Corresponding author.}
\renewcommand{\thefootnote}{\arabic{footnote}}

\input{01-intro}

\input{02-background}

\input{03-method}

\input{04-results}

\input{05-conclusion-discussion}

\newpage
\bibliography{main}
\newpage
\appendix
\end{document}

%% file: 00-abs.tex
This work introduces \textbf{Hybrid Sparse Attention (\aname)}, a new architecture that interleaves each full attention layer with several sparse attention layers. While conceptually simple, \aname strategically derives each sparse layer’s token selection and KV caches directly from the preceding full attention layer.
This architecture resolves two fundamental limitations of prior sparse attention methods.
First, conventional approaches typically rely on additional proxies to predict token importance, introducing extra complexity and potentially suboptimal performance. In contrast, \aname uses the full attention layer as a precise oracle to identify important tokens. Second, existing sparse attention designs often reduce computation without saving KV cache. \aname enables sparse attention layers to reuse the full attention KV cache, thereby reducing both computation and memory.
We evaluate \aname on both 7B dense and 80B MoE models. Across all settings, \aname consistently outperforms both full attention and hybrid SWA baselines. Notably, in the 80B MoE model with 49 total layers, only 5 layers employ full attention, yet \aname achieves substantial performance gains while reducing KV cache storage by nearly $10\times$.

%% file: 01-intro.tex
\section{Introduction}

The demand for long-context capabilities has become a cornerstone of modern Large Language Models (LLMs), driven by emerging paradigms such as test-time scaling~\citep{o1,r1} and agentic workflows~\citep{k2, anthropic_building_effective_agents_2024}.
Yet, the self-attention mechanism in standard Transformers scales quadratically with sequence length, causing computational latency and cost to grow prohibitively as context length increases.

Sparse attention offers a straightforward and effective solution to mitigate this quadratic bottleneck~\citep{sparsetransformer}.
Sparse attention computes attention over a selected subset of important tokens rather than all tokens in the sequence.
Existing methods can be broadly categorized into training-free and trainable approaches. Training-free methods rely on fixed patterns or heuristic to select important tokens~\citep{h2o,streamingllm,minference,quest,lserve,duo}. Trainable sparse attention learns which tokens to attend during training, either via low-cost self-distillation~\citep{seerattn_v1,dsa} or by being directly integrated into pre-training~\citep{nsa, moba}. Nevertheless, sparse attention architectures still suffer from two fundamental limitations:

\textbf{(1) Proxy-based Sparse Token Selection.}
Sparse attention fundamentally depends on a selection mechanism to identify important tokens prior to attention computation.
Existing methods typically rely on lightweight proxies, such as predefined patterns, heuristics, approximate estimates, or additional selection modules~\citep{minference,seerattn_v1, moba, seer-r, nsa, dsa}.
However, these proxies are inherently approximate and may fail to capture true token importance, particularly in long and evolving contexts.
As a result, sparse token selection is often bounded by the fidelity of the proxy, potentially limiting the expressive power of sparse attention.
While learnable sparse attention alleviates selection errors by learning token selection during training, it does not fundamentally eliminate the proxy-based bottleneck and introduces additional selection modules that increase training complexity.

\textbf{(2) Computation Reduction without Memory Relief.}
Modern sparse attention methods increasingly adopt dynamic sparsity to preserve model fidelity. Unlike static patterns (e.g., fixed strides or block structures), which can reduce KV cache storage but often incur noticeable performance degradation, dynamic approaches typically retain the full KV cache. This is because complete KV cache eviction is irreversible and destructive, as token importance may shift as generation progresses and context evolves. While dynamic sparse attention can effectively reduce computation, it provides no relief for memory consumption. Maintaining a full-sized KV cache therefore remains a dominant bottleneck for serving throughput and maximum batch size, limiting the practical benefits of sparse attention in long-context settings.

To address these challenges, we introduce \textbf{Hybrid Sparse Attention (\aname)}.
The key idea is to interleave every full attention layer with multiple sparse attention layers, where the sparse layers strategically derive important token selection and KV caches from the preceding full layer.
This design is motivated by two empirical observations in recent literature: token saliency is stable across consecutive layers (\S\ref{back:salient}), and cross-layer KV cache sharing reduces memory without hurting performance (\S\ref{back:kvsharing}).
In \aname, full attention can precisely identify important token selection and already produces KV caches, which sparse layers can directly reuse.
By reusing the important token indices from full attention, sparse selection becomes oracle-guided. 
This eliminates the need for auxiliary proxy modules and ensures stable end-to-end training. 
By reusing the KV caches from full attention, sparse attention adds no per-layer KV overhead, effectively alleviating the memory pressure associated with dynamic sparse attention.
Meanwhile, inspired by hybrid sliding window attention (SWA) architectures~\citep{gemma3,gpt-oss,xiao2026mimo}, \aname augments each sparse attention layer with an additional SWA branch that maintains a small, local KV cache to enhance short-range modeling capacity.

We evaluate \aname on both 7B dense and 80B Mixture-of-Experts (MoE) model settings. 
For the 7B dense model, we adopt a full-to-sparse layer ratio of 1:3, while for the 80B MoE model, a more aggressive 1:11 ratio is used. In both cases, the final layer employs full attention to preserve global aggregation. 
Across tasks and context lengths, \aname consistently outperforms both full attention and hybrid SWA baselines, without incurring any additional KV cache cost relative to the hybrid SWA baseline. 
Remarkably, in the \aname 80B MoE model with 49 total layers, \textit{only 5 layers use full attention}, meaning nearly $10\times$ KV cache reduction, while the models still delivers substantial performance gains.
Compared with Hybrid SWA, \aname can significantly reduce the number of full attention layers, effectively pushing the hybrid ratio to its limit.  
In summary, these results indicate that \aname provides a simple and effective architectural solution to the core limitations of sparse attention, achieving strong long-context modeling capability with clear efficiency and memory advantages.

%% file: 02-background.tex
\section{Background \& Motivation}

\subsection{Training-free vs. Trainable Sparse Attention}

Sparse attention methods can be divided into training-free and trainable approaches.
Training-free methods rely on fixed patterns or heuristics to identify important tokens. 
They are applied as a drop-in modification at inference, enabling fast sparsity decisions with minimal computational cost~\citep{streamingllm,quest,duo,h2o}. 
However, applying sparsity only at inference creates a training–inference mismatch, which may lead to error accumulation in long decoding or multi-step reasoning~\citep{hu2026lil, liu2025quantizationreasoning, he2025nondeterminism}.
Trainable sparse attention methods, in contrast, learn token importance during training through lightweight selection modules. 
By integrating sparsity into the training process, they improve alignment between training and inference, selecting more informative tokens with higher recall and overall accuracy~\citep{seerattn_v1, seer-r, dsa, nsa, moba, minicpm4, zhao2025infllm}. 
However, training these selection modules are non-trivial. 
One approach uses auxiliary losses, such as self-distillation, to align the gating or indexer module with the original dense attention~\citep{seerattn_v1, seer-r, dsa}. 
These methods are simple but suboptimal. 
Alternatively, NSA~\citep{nsa} performs end-to-end sparse pretraining by injecting the compressed attention (selection module) output into the main attention. 
This design allows the selection module to receive learning signals only indirectly through the final attention output, without direct supervision on its token selection decisions.

\subsection{Hybrid Attention Architecture}

To reduce quadratic compute and KV cache costs, hybrid attention has emerged as a promising solution for scaling context length.
For instance, MiniMax-01~\citep{li2025minimax} integrates both linear attention and softmax attention mechanisms in a structured pattern. Similarly, Qwen3-Next~\citep{qwen3next2025} and Kimi Linear~\citep{kda} incorporate Gated DeltaNet \citep{gdn} or its variants. The Nemotron family~\citep{nemotron2, blakeman2025nemotronh} and Jamba~\citep{lieber2024jamba} integrate Mamba modules \citep{mamba, mamba2} with standard self-attention modules. Models such as GPT-OSS~\citep{gpt-oss}, Gemma3~\citep{gemma3}, and MiMo-V2-Flash~\citep{xiao2026mimo} employ a heterogeneous interleaving of sliding window attention and global full attention layers. The sliding window size can be as small as 128 tokens with negligible KV cache overhead. Yet, the hybrid model with dynamic sparse attention has not been fully explored.

\subsection{Cross-Layer Salient Token Stability} \label{back:salient}

Several concurrent works have observed that salient tokens (sparse tokens with higher attention scores) tend to remain relatively stable across consecutive layers in standard transformer models~\citep{yang2024tidaldecode, hao2025omnikv, yang2025lessismore, zarch2025delta, deshmukh2025kascade}.
These methods exploit this property to accelerate inference as a training-free manner. Specifically, they identify important tokens using a full attention layer and reuse the salient token indices in subsequent layers to perform sparse attention computation.
Inspired by these works, we elevate this empirical observation to a hybrid attention architecture for pretraining, in which full attention layers identify important tokens that are subsequently reused by following sparse attention layers.

\subsection{Cross-layer KV Cache Sharing} \label{back:kvsharing}

Cross-layer KV cache sharing is a memory optimization technique in which Key and Value tensors computed in one layer are reused by subsequent layers, instead of being recomputed and stored independently for every layer. 
This design substantially reduces the KV cache memory footprint, while empirical studies show that it incurs little to no degradation in model accuracy.
YOCO~\citep{yoco}, CLA~\citep{cla}, the Apple Foundation Model~\citep{li2025apple}, and Gemma 3n~\citep{gemma3} integrate cross-layer KV cache sharing mechanisms directly into their model architectures. 
SwiftKV~\citep{qiao2025swiftkv} adapts standard pretrained models to support cross-layer KV cache sharing over a subset of layers via distillation. 
MiniCache~\citep{liu2024minicache} also observes that KV cache exhibits high similarity between adjacent layers in the middle-to-deep regions of LLMs, and proposes cross-layer compression methods to exploit this redundancy.

%% file: 03-method.tex
\section{Methodology}
\subsection{\aname Overview}

\begin{figure*}[h]
    \centering
    \includegraphics[width=1\linewidth]{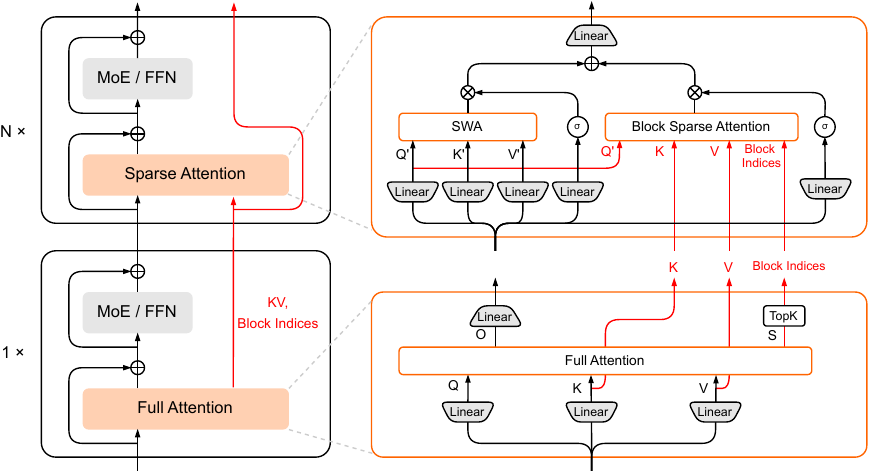}
    \caption{\aname Architecture Diagram. Each full attention layers is interleaved with multiple sparse attention layers. Sparse attention directly reuses the KV cache and important token indices from the preceding full attention layer.}
    \label{fig:architecture}
\end{figure*}

As shown in Figure~\ref{fig:architecture}, \aname architecture replaces the standard Transformer backbone with repeated hybrid blocks that are composed of one full attention layer followed by $N$ consecutive sparse attention layers.
At its core, both the sparse important token indices and the KV caches used in these sparse layers are directly derived from the preceding full attention layer within the same block.

The full attention layer computes standard scaled dot product self-attention but additionally outputs \textit{block-wise attention importance scores} $S$, from which we derive \textit{TopK block indices}. These indices are then reused by the next $N$ sparse layers. To reduce KV cache memory and bandwidth consumption, \aname further incorporates cross-layer KV cache sharing. The sparse attention layers reuse the KV cache produced by the preceding full attention layer within a hybrid block for the \textit{block sparse attention branch}. The sliding window attention (SWA) branch, in contrast, maintains its own lightweight KV cache to enhance short-range modeling capacity. Finally, sigmoid gates are applied to the output of the two branches~\citep{qiu2025gated} before summed as the final attention output. 

% M x (1 full attention + N x (spasre attention(kv from full attention) + swa (kv its own)))

\subsection{Full Attention Layers}

The full attention layers follow the standard softmax self-attention formulation used in Transformers. To identify salient tokens for subsequent sparse attention, the attention scores must be exposed for selection. However, materializing the full attention matrix is prohibitively expensive in terms of both memory and bandwidth. Consequently, modern Transformers rely on FlashAttention~\citep{flash1, flash2, flash3}, which avoids explicitly storing attention scores by computing softmax in a tiled manner with online normalization.

To mitigate this issue, instead of outputting the full attention score matrix, \aname only materializes block-level (tile-level) maximum attention scores for TopK selection. With the standard self-attention formulation:
\begin{align}
\boldsymbol{q}_t,\ \boldsymbol{k}_t,\ \boldsymbol{v}_t
= \mathbf{W}_{q/k/v}\,\boldsymbol{x}_t .
\end{align}
\begin{align}
\boldsymbol{o}_t
= \sum_{i=1}^{t}
\frac{
\exp\!\left(\frac{\boldsymbol{q}_t^{\top}\boldsymbol{k}_i}{\sqrt{d}}\right)
}{
\sum_{j=1}^{t}\exp\!\left(\frac{\boldsymbol{q}_t^{\top}\boldsymbol{k}_j}{\sqrt{d}}\right)
}
\,\boldsymbol{v}_i ,
\end{align}
where $d$ is the head dimension size. Let $B$ be the block size of our attention score output, and there will be $\lceil t/B\rceil$ number of blocks. We define the column token index set at block index $i$ as
$\mathcal{B}_i=\{(i-1)B+1,\ldots,\min(iB,N)\}$. Then the block-level max attention score $\mathbf{S}\in R^{ t\times  \lceil t/B\rceil }$ will be:
\begin{align}
\mathbf{S}_{t}^{i}
= \max_{i^{'}\in\mathcal{B}_i}\ 
\left(\frac{
\exp\!\left(\frac{\boldsymbol{q}_t^{\top}\boldsymbol{k}_{i^{'}}}{\sqrt{d}}\right)
}{
\sum_{j=1}^{t}\exp\!\left(\frac{\boldsymbol{q}_t^{\top}\boldsymbol{k}_j}{\sqrt{d}}\right)
}\right)
\end{align}

We find that the $\mathbf{S}_t$ can be easily obtained by slightly modifying the FlashAttention kernel, following the approach similar to~\citep{seerattn_v1, seer-r}. Specifically, FlashAttention already computes the row-wise maximum of the attention logits during its online softmax procedure, and this intermediate result can be reused to derive block-wise attention scores by storing and appropriately rescaling it.

\begin{algorithm}[]
% \small
\caption{\textsc{FlashAttention} with Block Attention Score Output (assuming $B = B_N$ for simplicity)}
\label{alg:oracle_attn}
\begin{algorithmic}[1]
\Require Queries $\mathbf{Q} \in \mathbb{R}^{t \times d}$, Keys $\mathbf{K} \in \mathbb{R}^{t \times d}$, Values $\mathbf{V} \in \mathbb{R}^{t \times d}$, softmax scale $\tau$
\Ensure Attention output $\mathbf{O} \in \mathbb{R}^{t \times d}$, Block attention scores $\mathbf{S} \in \mathbb{R}^{t \times \lceil t / B \rceil}$
\State $T_r \gets \lceil t / B_M \rceil$, \quad $T_c \gets \lceil t / B_N \rceil$
\State \Initialize $\mathbf{O}_i$, $\mathbf{S}_i$, $\boldsymbol{m}_i$, $\boldsymbol{\ell}_i$
\For{$i = 0, \ldots, T_r - 1$}
    \State Load $\mathbf{Q}_i \in \mathbb{R}^{B_M \times d}$ from HBM to SRAM
    \For{$j = 0, \ldots, T_c - 1$}
        \State Load $\mathbf{K}_j, \mathbf{V}_j \in \mathbb{R}^{B_N \times d}$ from HBM to SRAM
        \State $\mathbf{A}_{ij} \gets \mathbf{Q}_i \mathbf{K}_j^\top \cdot \tau$
        \State $\tilde{\boldsymbol{m}}_{ij} \gets \text{rowmax}(\mathbf{A}_{ij})$, Store $\tilde{\boldsymbol{m}}_{ij}$ to $\mathbf{S}_{ij}$
        \State $\boldsymbol{m}_i \gets \text{max}(\boldsymbol{m}_i, \tilde{\boldsymbol{m}}_{ij})$
        \State Update $\mathbf{O}_i, \boldsymbol{\ell}_i$ as in \textsc{FlashAttention}
    \EndFor
    \State $\mathbf{O}_i \gets \mathbf{O}_i / \boldsymbol{\ell}_i$, Write $\mathbf{O}_i$ to HBM
    \For{$j = 0, \ldots, T_c - 1$}
        \State $\mathbf{S}_{ij} \gets (\mathbf{S}_{ij} - \boldsymbol{m}_i) \,/\, \boldsymbol{\ell}_i$, Write $\mathbf{S}_{ij}$ to HBM
    \EndFor
\EndFor
\State \Return $\mathbf{O}, \mathbf{S}$
\end{algorithmic}
\end{algorithm}

Algorithm~\ref{alg:oracle_attn} summarizes the modified FlashAttention procedure, and we assume the sparse attention block size $B$ is the same as $B_N$ for simplicity. In addition to the standard attention output, the kernel emits block-level attention scores that can be directly used for salient block selection in later sparse attention layers with negligible overhead. 

With the block-wise attention scores $S$, we apply a TopK operator to select key-block indices $\mathcal{I}$ that are reused by the subsequent sparse attention layers. Noted that attending to $k$ tokens in sparse attention corresponding to selecting $k/B$ TopK blocks of tokens, where $B$ is the block size. In \aname, the default $k$ and $B$ is 1024 and 64, respectively.
Under Grouped-Query Attention (GQA)~\citep{gqa}, we further aggregate $S$ within each query group (via a group-wise maximum) so that all heads in the same group share identical sparse indices, improving sparse attention kernel efficiency and reducing indexing overhead.

\subsection{Sparse Attention Layers}
Each sparse layer contains two attention branches that operate on the \textit{same query} but use \textit{different KV} sources.
\textit{Block Sparse Attention branch} attends only to key-value blocks indexed by $\mathcal{I}$, where both $\mathcal{I}$ and the KV cache are derived from the preceding full attention layer.
\textit{SWA branch} attends to a local sliding window of size $w$ with its own small KV cache, improving locality and expressivity. $w$ is set to be 128 in \aname implementation.
The two branch outputs are then fused via a lightweight sigmoid gate. The detailed processes can be model as below. First, we compute the standard SWA branch output using:
\begin{align}
&\boldsymbol{q}_t', \boldsymbol{k}_t', \boldsymbol{v}_t' 
= \mathbf{W}_{q'/k'/v'} \boldsymbol{x}_t \\
\boldsymbol{o}_t'
= \sum_{i = t-w+1}^{t}&
\frac{
\exp\!\left(\frac{\boldsymbol{q}_t'^{\top} \boldsymbol{k}'_i}{\sqrt{d}}\right)
}{
\sum_{j = t-w+1}^{t} \exp\!\left(\frac{\boldsymbol{q}_t'^{\top} \boldsymbol{k}'_j}{\sqrt{d}}\right)
}
\, \boldsymbol{v}_i'
\end{align}

Then, when computing the sparse attention branch, we concatenate the selected key and value blocks from the shared $\mathbf{K}$, $\mathbf{V}$ from the full attention layers using block indices $\mathcal{I}$. The sparse attention and SWA branch uses the same query $\boldsymbol{q}_{t}'$, and the output can then be written as:
\begin{align}
\tilde{\mathbf{K}},\tilde{\mathbf{V}}
&=\mathrm{concat}\Big(\{\mathbf{K}/\mathbf{V}_{[(j-1)B+1:\,jB]}\}_{j\in\mathcal{I}}\Big)\\
\tilde{\boldsymbol{o}}_t
&= \sum_{i=1}^{k}
\frac{
\exp\!\left(\frac{\boldsymbol{q}_t'^{\top} \tilde{\boldsymbol{k}}_{i}}{\sqrt{d}}\right)
}{
\sum_{j=1}^{k} \exp\!\left(\frac{\boldsymbol{q}_t'^{\top}
\tilde{\boldsymbol{k}}_{j}}{\sqrt{d}}\right)
}
\, \tilde{\boldsymbol{v}}_{i} 
\end{align}

Finally, we apply a sigmoid gate on both branch outputs and sum them to obtain the final attention layer output~\citep{qiu2025gated}. 
\begin{align}
\tilde{g}_t, g_{t}'&= \sigma\!\left(\mathbf{W}_{\tilde{g}/g'} \boldsymbol{x}_t\right)\\
\boldsymbol{o}_t &= \tilde{g}_t \odot \tilde{\boldsymbol{o}}_t \;+\; g_t'\odot \boldsymbol{o}_t'
\end{align}

Through our experiments, we find that maintaining an independent KV cache for the SWA branch is essential for preserving model expressivity. One possible explanation is that SWA primarily serves as a local information pathway and requires different representations to capture short-range coherence, whereas the KV shared from the preceding full attention layer is optimized for global retrieval and may lack sufficient local features. The two-branch gated fusion yields a dynamic mixture of global and local information while remaining efficient in both computation and memory usage.

%% file: 04-results.tex
\section{Experiments}

\subsection{Experiments Setup}

\begin{table}[h]
\small
\centering
\begin{tabular}{lcc}
\toprule
\textbf{Configuration} & \textbf{7B Dense} & \textbf{80B MoE} \\
\midrule
Layers                          & $36$     & $49$ \\
Attention Heads (Q / KV)        & $32 / 8$  & $64/4$ \\
Head Dimensions                 & $128$    & $128$ \\
Hidden Size                     & $4096$   & $2048$ \\
Hybrid Ratio (Full : Sparse)    & $1:3$    & $1:11$ \\
Sliding Window Size             & $128$    & $128$ \\
Sparse Attn Block Size          & $64$     & $64$ \\
Sparse Attn TopK Tokens      & $1024$   & $1024$ \\
\midrule
MoE Expert (Activated/Total) & -- & $8:512$ \\
\bottomrule
\end{tabular}
\caption{Model Architecture Configurations.}
\label{tab:modelconfig}
\end{table}

\paragraph{Model Configuration}
In the following experiments, we use a standard 7B dense Transformer with 36 layers and an 80B-A3B MoE model with 49 layers. 
We adopt Grouped-Query Attention (GQA), using 32 query heads with 8 KV heads for the 7B dense model, and 64 query heads with 4 KV heads for the 80B MoE model. 
We evaluate three architectures: 
(1) \textbf{Full-Attn}: all layers use standard full attention. 
(2) \textbf{Hybrid SWA}: hybrid sliding window attention models, using a full-to-SWA layer ratio of 1:3 for the 7B model and 1:11 for the 80B MoE model. 
(3) \textbf{\aname}: hybrid models with the same hybrid ratios as Hybrid SWA, but augmenting SWA with the proposed sparse attention (Top-1024 tokens with block size 64). 
For all hybrid models, the final layer uses full attention. For sparse attention and sliding window attention, we incorporate per-head learnable sink biases, following the approach in gpt-oss~\citep{gpt-oss}. 
For Full-Attn in the MoE setting, we additionally employ gated attention~\citep{qiu2025gated} to stabilize training.
Detailed model configurations are listed in Table~\ref{tab:modelconfig}.

\paragraph{Training Hyper-parameters}  
For the 7B models, we first train on 1T tokens with a sequence length of 8,192 using the AdamW optimizer 
($\beta_1=0.9$, $\beta_2=0.95$, $\epsilon=10^{-10}$), weight decay 0.1, and gradient clipping with a maximum norm of 1.0. 
Training uses BF16 precision and a WSD schedule with a maximum learning rate of $8.3 \times 10^{-4}$. 
To extend the models for long-context evaluation, we further train on 200B tokens with a sequence length of 32,768 and a learning rate of $3.0 \times 10^{-5}$. The RoPE base frequency is adjusted to 640,000 at this stage.
For the 80B MoE model, training is performed on 500B tokens with a sequence length of 32,768 using the WSD schedule with a maximum learning rate of $1 \times 10^{-3}$, and the RoPE base frequency is also set to 640,000.

\paragraph{Evaluation Benchmark}
We evaluate \aname based on a series of benchmarks, encompassing various capabilities: (1) General language understanding and reasoning, including BBH~\citep{suzgun2022challenging}, MMLU~\citep{hendrycks2020measuring}, MMLU-Redux~\citep{gema2024we}, MMLU-Pro~\citep{wang2024mmlu}, DROP~\citep{dua2019drop}, ARC~\citep{clark2018think}, HellaSwag~\citep{zellers2019hellaswag}, WinoGrande~\citep{sakaguchi2021winogrande}, TriviaQA~\citep{joshi2017triviaqa},
(2) Mathematics reasoning, including
GSM8K~\citep{cobbe2021training}, MATH~\citep{hendrycks2021measuring}
(3) Coding, including HumanEval~\citep{humaneval}, MBPP~\citep{mbpp}, 
(4) Chinese understanding, including C-Eval~\citep{huang2023c}, CMMLU~\citep{li2023cmmlu}.
(5) Long context, Ruler~\citep{hsieh2024ruler}.

\subsection{Performance of \aname on General Benchmarks}

\begin{table*}[h]
    \centering
    \small
    \resizebox{0.9\textwidth}{!}{
    \begin{tabular}{l c | c c c | c c c}
    \toprule
    \multirow{2}{*}{\textbf{Benchmark}} & \multirow{2}{*}{\textbf{\# Shots}} &
    \multicolumn{3}{c|}{\textbf{7B Dense (Hybrid 1:3)}} &
    \multicolumn{3}{c}{\textbf{80B MoE (Hybrid 1:11)}} \\
    \cmidrule(lr){3-5} \cmidrule(lr){6-8}
    & &
    \textbf{Full-Attn} & \textbf{Hybrid SWA} & \textbf{\aname} &
    \textbf{Full-Attn} & \textbf{Hybrid SWA} & \textbf{\aname} \\
    \midrule

    \multicolumn{8}{l}{\textbf{General}} \\
    BBH & 3-shot & 52.2 & \textbf{54.0} & 53.5 & 56.1 & 48.2 & \textbf{56.3} \\
    MMLU & 5-shot & 56.9 & 57.5 & \textbf{58.8} & 61.8 & 54.9 & \textbf{62.2} \\
    MMLU-Redux & 5-shot & 59.6 & 60.8 & \textbf{61.6} & 65.6 & 57.4 & \textbf{66.2} \\
    MMLU-Pro & 5-shot & 26.8 & 26.5 & \textbf{29.0} & \textbf{33.8} & 27.2 & {32.6} \\
    DROP & 3-shot & \textbf{53.1} & 43.8 & 52.4 & \textbf{56.7} & 47.8 & {56.5} \\
    ARC-Challenge & 25-shot & 70.2 & 74.9 & \textbf{75.0} & \textbf{78.4} & 63.9 & {77.6} \\
    HellaSwag & 10-shot & 77.5 & 77.8 & \textbf{78.1} & 78.2 & 77.1 & \textbf{79.0} \\
    WinoGrande & 5-shot & 73.7 & \textbf{74.9} & 74.3 & 71.2 & 69.0 & \textbf{72.1} \\
    TriviaQA & 5-shot & 50.1 & 50.0 & \textbf{51.1} & 54.7 & 52.2 & \textbf{55.5} \\
    \midrule

    \multicolumn{8}{l}{\textbf{Mathematics}} \\
    GSM8K & 8-shot & 33.3 & 35.6 & \textbf{37.9} & 53.8 & 45.3 & \textbf{54.1} \\
    MATH & 4-shot & 9.2 & 9.2 & \textbf{10.1} & 28.6 & 25.8 & \textbf{30.8} \\
    \midrule

    \multicolumn{8}{l}{\textbf{Code}} \\
    HumanEval & 0-shot & \textbf{25.0} & 22.0 & 23.5 & 35.4 & 31.7 & \textbf{38.4} \\
    MBPP & 3-shot & 51.0 & \textbf{52.8} & 51.6 & 55.3 & 51.9 & \textbf{59.3} \\
    \midrule

    \multicolumn{8}{l}{\textbf{Chinese}} \\
    C-Eval & 5-shot & 50.6 & 50.6 & \textbf{52.2} & 64.6 & 58.8 & \textbf{65.0} \\
    CMMLU & 5-shot & 52.5 & 52.9 & \textbf{54.5} & 66.7 & 58.4 & \textbf{67.0} \\
    \bottomrule
    \end{tabular}
    }
    \caption{
        Comparison of \aname across 7B dense models (trained on 1T tokens) and 80B MoE models (trained on 500B tokens) with Full-Attn and Hybrid SWA. Best results in each row are highlighted in bold.
    }
    \label{tab:7b_eval}
\end{table*}

\paragraph{7B Dense Performance}
Table~\ref{tab:7b_eval} compares 7B models under three attention variants. 
Overall, \aname achieves strong performance across general benchmarks, mathematics, and Chinese understanding, while Hybrid SWA offers a competitive and computationally efficient baseline that is particularly strong on BBH and MBPP+. 
Specifically, \aname surpasses the Full-Attn baseline on a broad set of knowledge and reasoning benchmarks, including MMLU (58.8 vs.\ 56.9), MMLU-Redux (61.6 vs.\ 59.6), and MMLU-Pro (29.0 vs.\ 26.8), suggesting that sparse token selection can preserve (and even enhance) global reasoning and factual recall despite reduced attention computation. 
\aname also yields consistent gains on multi-step reasoning tasks such as GSM8K and MATH.
On classic multiple-choice commonsense and reading comprehension benchmarks, \aname is either best or comparable, with slight improvements on ARC-Challenge, HellaSwag, and TriviaQA. 
For Chinese benchmarks, \aname provides clear gains on both C-Eval (52.2 vs.\ 50.6) and CMMLU (54.5 vs.\ 52.5).

\paragraph{Scaling to 80B MoE}
We further evaluate \aname in an 80B-A3B MoE setting with a more aggressive full-to-sparse layer ratio of 1:11. 
Despite having only five full attention layers, \aname outperforms both Full-Attn and Hybrid SWA across nearly all benchmarks; only MMLU-Pro, DROP, and ARC-C are slightly lower than Full-Attn. 
Notably, the performance gains are often larger than those observed in the 7B dense setting.
In this regime, Hybrid SWA exhibits noticeable accuracy degradation compared to Full-Attn on several benchmarks, including BBH, DROP, the MMLU series, GSM8K, and Chinese understanding. 
This suggests that relying solely on local window attention becomes insufficient as the hybrid ratio becomes more aggressive. 
By introducing the sparse attention branch, \aname mitigates this gap by recovering access to globally relevant tokens selected from the full attention, and in many cases even surpasses the Full-Attn baseline, while requiring $10\times$ less KV cache.
\textit{These results highlight a key advantage of \aname: the number of full attention layers can be substantially reduced without sacrificing modeling capability, which is also the design rationale of \aname.}

\subsection{Long-context Benchmarks}

\begin{table}[h]
\centering
\resizebox{\textwidth}{!}{%
\begin{tabular}{lllcccccccccccc}
\toprule
\textbf{Size} & \textbf{Ctx} & \textbf{Type} & \textbf{S1} & \textbf{S2} & \textbf{S3} & \textbf{MK1} & \textbf{MK2} & \textbf{MK3} & \textbf{MQ} & \textbf{MV} & \textbf{VT} & \textbf{CWE} & \textbf{FWE} & \textbf{Total} \\
\midrule

% 7B Section
\multirow{6}{*}{\textbf{7B}} 
 & \multirow{3}{*}{\textbf{16k}} 
   & Full-Attn  & \textbf{100.0} & \textbf{100.0} & \textbf{100.0} & 98.6 & \textbf{100.0} & 96.4 & \textbf{99.2} & \textbf{99.4} & 94.4 & 37.1 & 97.4 & 93.0 \\
 & & Hybrid SWA & \textbf{100.0} & \textbf{100.0} & 99.8 & 97.6 & 99.8 & \textbf{96.6} & 97.4 & 94.7 & \textbf{99.1} & 23.7 & \textbf{98.5} & 91.6 \\
 & & \aname      & \textbf{100.0} & \textbf{100.0} & \textbf{100.0} & \textbf{99.4} & 99.6 & \textbf{96.6} & 96.7 & 89.8 & 97.2 & \textbf{60.8} & 95.5 & \textbf{94.1} \\
 \cmidrule{2-15}
 & \multirow{3}{*}{\textbf{32k}} 
   & Full-Attn  & \textbf{100.0} & \textbf{100.0} & \textbf{100.0} & 98.0 & {99.4} & 75.8 & \textbf{96.4} & \textbf{98.3} & 90.5 & 16.6 & \textbf{95.1} & 88.2 \\
 & & Hybrid SWA & \textbf{100.0} & \textbf{100.0} & 99.6 & 96.2 & 98.8 & 53.4 & 93.4 & 87.2 & \textbf{98.5} & 10.4 & 88.5 & 84.2 \\
 & & \aname      & \textbf{100.0} & \textbf{100.0} & 99.8 & \textbf{98.2} & \textbf{99.6} & \textbf{76.2} & 88.8 & 94.7 & 91.2 & \textbf{38.8} & \textbf{95.1} & \textbf{89.3} \\
\midrule

% 80B Section
\multirow{6}{*}{\textbf{80B}} 
 & \multirow{3}{*}{\textbf{16k}} 
   & Full-Attn  & 100.0 & \textbf{99.8} & 92.6 & \textbf{99.6} & 99.2 & 93.0 & \textbf{97.3} & \textbf{94.9} & \textbf{95.4} & \textbf{74.5} & 80.4 & \textbf{93.6} \\
 & & Hybrid SWA & 95.2 & 94.8 & 70.8 & 93.2 & 86.4 & 69.4 & 83.2 & 57.8 & 66.7 & 13.3 & 69.2 & 72.7 \\
 & & \aname     & \textbf{100.0} & \textbf{99.8} & \textbf{99.0} & {98.2} & \textbf{100.0} & \textbf{99.6} & 92.2 & {91.3} & 90.3 & {40.2} & \textbf{86.4} & {90.6} \\
 \cmidrule{2-15}
 & \multirow{3}{*}{\textbf{32k}} 
   & Full-Attn  & \textbf{100.0} & 99.2 & 81.2 & \textbf{99.0} & \textbf{99.4} & 77.0 & {86.0} & 79.5 & 74.5 & \textbf{40.7} & 66.7 & 82.1 \\
 & & Hybrid SWA & \textbf{100.0} & 98.2 & 58.8 & 89.6 & 90.2 & 61.0 & 74.1 & 56.6 & 73.1 & 8.4 & 54.3 & 69.5 \\
 & & \aname      & \textbf{100.0} & \textbf{100.0} & \textbf{99.4} & 96.8 & {99.0} & \textbf{98.4} & \textbf{89.5} & \textbf{85.7} & \textbf{89.6} & {20.8} & \textbf{82.1} & \textbf{87.4} \\
\bottomrule
\end{tabular}%
}
\caption{RULER benchmark performance for 7B dense and 80B MoE models. 
Note that all 7B models were first trained on 1T tokens with a sequence length of 8K, and then further trained on 200B tokens with a sequence length of 32K. 
All 80B MoE models were trained on 500B tokens with a sequence length of 32K.
}
\label{tab:ruler}
\end{table}

Table \ref{tab:ruler} shows that \aname consistently preserves strong long-context accuracy. 
For the 7B dense model, \aname improves the overall score over both baselines at 16k and 32k, reaching totals of 94.1 and 89.3 (vs. 93.0 and 88.2 for Full-Attn and 91.6 and 84.2 for Hybrid SWA). The gains are most apparent on the harder multi-key/value and reasoning-heavy subsets, e.g., \aname substantially boosts CWE compared to baselines (60.8 at 16k and 38.8 at 32k), indicating better robustness as context grows. 
For the 80B MoE model, Hybrid SWA degrades sharply (72.7 at 16k and 69.5 at 32k) under an aggressive hybrid ratio, whereas \aname remains competitive with Full-Attn at 16k (90.6 vs. 93.6) and notably surpasses it at 32k (87.4 vs. 82.1), driven by large recoveries on difficult components such as MK3 (98.4 vs. 77.0) and stronger VT/MQ/MV stability. In general, \aname provides on-par long-context capabilities compared with Full-Attn across settings, while significantly reducing computation and KV cache size.

\subsection{Ablation Study}

\begin{figure*}[h]
    \centering
    \includegraphics[width=1\linewidth]{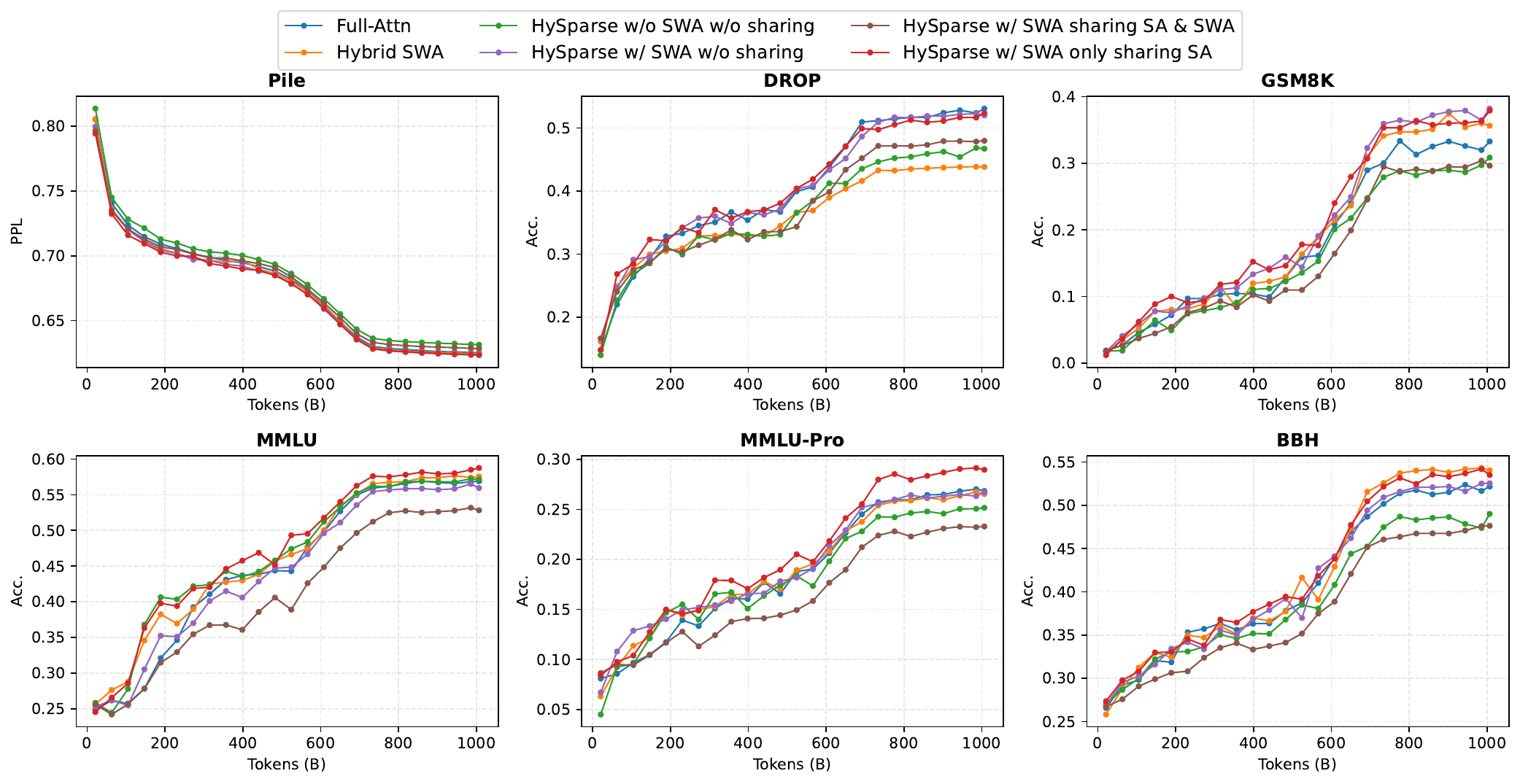}
    \caption{\aname Accuracy vs. Training Iterations.}
    \label{fig:acc vs iter}
\end{figure*}

In this section, we present detailed ablation studies on key architecture design choices, focusing on: (1) whether to include an intra-layer SWA branch within each sparse layer, and (2) how KV cache sharing is applied across the sparse attention and SWA branches. All ablation studies are conducted on the 7B dense models. The results are summarized in Table~\ref{tab:ablation} and Figure~\ref{fig:acc vs iter}.

\begin{table*}[h]
\centering
\small
\setlength{\tabcolsep}{3pt}
\renewcommand{\arraystretch}{1.3}

\begin{tabularx}{\linewidth}{l X *{5}{c}}
\toprule
\textbf{Group} & \textbf{Method} & \textbf{DROP} & \textbf{GSM8K} & \textbf{MMLU} & \textbf{MMLU-Pro} & \textbf{BBH} \\
\midrule

\multirow{2}{*}{\makecell[l]{Baselines}}
& Full-Attn   & 52.6 & 32.6 & 56.8 & 26.8 & 52.1 \\
& Hybrid SWA  & 43.9 & 35.7 & 57.6 & 26.6 & 54.3 \\
\midrule

\multirow{2}{*}{\makecell[l]{Oracle Token Selection\\(w/o KV cache sharing)}}
& \aname (w/o intra-layer-SWA) & 46.4 & 29.7 & \textbf{57.1} & 25.0 & 48.2 \\
& \aname (w/ intra-layer-SWA)   & \textbf{52.2} & \textbf{37.7} & 56.1 & \textbf{26.5} & \textbf{52.4} \\
\midrule

\multirow{2}{*}{\makecell[l]{KV Cache Sharing\\(w/ intra-layer SWA)}}
& \aname (sharing for SA \& SWA) & 47.9 & 30.2 & 52.8 & 23.2 & 47.2 \\
& \textbf{\aname (sharing only for SA)}        & \textbf{51.9} & \textbf{36.7} & \textbf{58.4} & \textbf{29.0} & \textbf{53.9} \\
\bottomrule
\end{tabularx}
\caption{Ablation of Different Architecture Design Choices on 7B experiments.}
\label{tab:ablation}
\end{table*}

\paragraph{Study 1: Intra-layer Hybridization with SWA}  
This study investigates whether sparse attention layers still benefit from an additional SWA branch when the sparse indices are provided by oracle token selection (with KV cache sharing disabled). A natural hypothesis is that the SWA branch might be redundant: if recent or local-context tokens are important, oracle selection from the preceding full attention layer should already include them among the selected blocks. However, our results indicate that removing the SWA branch leads to a clear accuracy drop. Specifically, we compare \textbf{\aname (w/o intra-layer SWA)} against \textbf{\aname (w/ intra-layer SWA)} under the same oracle-selected sparse indices, with both branches sharing the same QKV projection layers.

As shown in Table~\ref{tab:ablation}, adding the intra-layer SWA branch yields consistent improvements across most benchmarks: DROP increases from 46.4 to 52.2 (+5.8), GSM8K from 29.7 to 37.7 (+8.0), MMLU-Pro from 25.0 to 26.5 (+1.5), and BBH from 48.2 to 52.4 (+4.2). These gains suggest that even with high-quality sparse selection, a dedicated sliding window pathway remains important for modeling short-range coherence and local computation patterns that are not reliably captured by sparse global retrieval alone. Additionally, the SWA branch may help stabilize optimization by providing a consistent local pathway, particularly during early training stages.

\paragraph{Study 2: Cross-layer KV Cache Sharing Configuration}
This study examines how KV cache sharing should be applied when each sparse layer contains both a sparse attention (SA) branch for global retrieval and a sliding window attention (SWA) branch for local modeling. 
A natural design choice is to maximize memory reuse by sharing the same KV cache for both branches, i.e., reusing the KV cache produced by the preceding full attention layer for both SA and SWA. 
However, this coupling can be overly restrictive because the two branches serve distinct roles: SA primarily requires globally informative keys and values aligned with block-level retrieval, whereas SWA benefits from a dedicated local representation that emphasizes short-range coherence and local computation patterns. 
In our KV cache sharing experiments, we compare \textbf{\aname (sharing for both SA \& SWA)} against \textbf{\aname (sharing only for SA)}.

As shown in Table~\ref{tab:ablation}, sharing KV caches for both SA and SWA substantially degrades accuracy. In contrast, sharing KV only for the SA branch while maintaining an independent KV cache for SWA recovers and improves performance across all evaluated tasks: DROP increases from 47.9 to 51.9 (+4.0), GSM8K from 30.2 to 36.7 (+6.5), MMLU from 52.8 to 58.4 (+5.6), MMLU-Pro from 23.2 to 29.0 (+5.8), and BBH from 47.2 to 53.9 (+6.7). These results suggest that SA can safely reuse the cross-layer KV cache from full attention to save GPU memory, whereas SWA should maintain its own KV cache to preserve strong local information. Forcing SWA to reuse the KV cache from the preceding full attention layer likely deprives it of the short-range, local features it requires and entangles it with globally optimized representations, thereby weakening the local pathway and reducing overall accuracy.

%% file: 05-conclusion-discussion.tex
\section{Discussion \& Future Works}

\paragraph{Can We Ultimately Avoid Full Attention?}  
Our findings connect to a broader trend in efficient attention, including hybrid attention and sparse attention methods. A recurring theme is that it remains challenging to \textit{completely} eliminate $O(n^2)$-style full attention components in practice: hybrid models retain explicit full attention layers, while sparse attention methods such as SeerAttention~\citep{seerattn_v1} and DSA~\citep{dsa} typically rely on gating or indexing mechanisms that still operate in $O(n^2)$, albeit in a compressed form. In this context, what matters most is the \emph{ratio} of expensive global computation to cheaper local or sparse computation, as well as GPU memory usage. Our results suggest that \aname, with oracle token selection and cross-layer KV sharing, provides a promising approach to reduce this ratio while preserving long-context modeling capabilities.

\paragraph{Potential of \aname for Efficient KV Cache Offloading}
\aname also points to a straightforward systems-level strategy for long-context serving: offload the full attention KV cache to external memory and pre-fetch it before computation, while keeping only the persistent selected/sparse KV on the GPU for subsequent sparse attention layers. Previous work such as OmniKV~\citep{hao2025omnikv} has explored similar approaches in a post-training setting. This technique has the potential to significantly reduce the KV cache footprint on GPU, enabling larger batch sizes and improving overall inference efficiency.

\section{Conclusion}

In this work, we introduced \textbf{Hybrid Sparse Attention (\aname)}, a simple yet effective hybrid attention architecture that interleaves each full attention layer with multiple sparse-attention layers.  
By strategically deriving important token selections and KV caches from preceding full attention layers, \aname eliminates the need for proxy-based token selection and enables sparse layers to operate without additional memory overhead.  
Importantly, \aname allows for a substantial reduction in the number of full attention layers in hybrid models without compromising modeling capabilities.  
In future work, we plan to scale \aname to even larger model sizes and train on more tokens to fully exploit its potential for efficient and accurate long-context modeling.